\documentclass{article}
\usepackage{spconf,amsmath,graphicx}

\usepackage{enumitem}
\setlist{nosep, leftmargin=14pt}

\usepackage{mwe} 

\title{VISION TRANSFORMERS FOR CLASSIFICATION OF BREAST ULTRASOUND IMAGES}
\name{Behnaz Gheflati and Hassan Rivaz}
\address{Department of Electrical and Computer Engineering, Concordia University, Canada}

\begin{document}
%
\maketitle
\begin{abstract}
Medical ultrasound (US) imaging has become a prominent modality for breast cancer imaging due to its ease-of-use, low-cost and safety. In the past decade, convolutional neural networks (CNNs) have emerged as the method of choice in vision applications and have shown excellent potential in automatic classification of US images. Despite their success, their restricted local receptive field limits their ability to learn global context information. Recently, Vision Transformer (ViT) designs that are based on self-attention between image patches have shown great potential to be an alternative to CNNs. In this study, for the first time, we utilize ViT to classify breast US images using different augmentation strategies. The results are provided as classification accuracy and Area Under the Curve (AUC) metrics, and the performance is compared with the state-of-the-art CNNs. The results indicate that the ViT models have comparable efficiency with or even better than the CNNs in classification of US breast images.

\end{abstract}
\section{Introduction}
\label{sec:intro}

Breast cancer is the most frequent cause of cancer mortality among women, making annual breast cancer screening essential for early detection and reducing the death rate. In recent years, ultrasound (US) imaging has become one of the most promising modalities due to its availability, real-time display, cost-efficiency, and non-invasive nature. Over the previous decades, US imaging has shown great potential in automated breast lesion classification, segmentation, and detection tasks. Automated analysis of medical images will help the radiologist to detect masses in US images and reduce the number of false negative readings in a cost-efficient way \cite{al2019deep, karimi2021convolution}.

Deep learning has lately become a leading tool in various research domains. Convolutional Neural Networks (CNNs) have been the most common networks for automatic medical image analysis applications such as image classification in recent years. However,  due to their localized receptive fields, these models have a poor performance in learning the long-range information, limiting their capabilities for vision tasks \cite{hatamizadeh2021unetr}.

The Transformer architecture, provided by Vaswani \textit{et al.} \cite{vaswani2017attention} is currently the dominant model in the field of natural language processing (NLP). Motivated by the success of the self-attention based deep neural networks of Transformer models in NLP, Dosovitskiy \textit{et al.} \cite{dosovitskiy2020image} introduced the Vision Transformer (ViT) architecture for the image classification application. The overall training process in these models is based on splitting the input image into patches and treating each embedded patch as a word in NLP. These models use self-attention modules to learn the relation between these embedded patches \cite{hatamizadeh2021unetr}. 

Herein, we take a step forward in exploiting Transformers in US medical image analysis and explore the potential application of self-attention based architectures to classify breast US images. In particular, we compared different pre-trained ViT models with different sizes and configurations based on their performance when fine-tuning for a downstream task. We then transfer these models for our specific task based on the US dataset. In addition, we also fine-tune our dataset based on the state-of-the-art (SOTA) CNNs. The results show the great potential of ViT models in breast US image classification. To the best of our knowledge, this is the first study to investigate the performance of ViT architectures on the classification of US breast images.

\section{RELATED WORK}
\label{sec:format}

\subsection{CNN-based classification networks}
\label{ssec:subhead}

During the past decade, CNNs have been used as a standard technique for medical image classification applications. Several studies have been conducted to benefit from the freely available CNN models pre-trained on a large amount of training data and transferring the pre-trained models for US image classification \cite{lazo2020comparison, tehrani2020pilot}.
In a comparative study conducted by Hatamizadeh \textit{et al.}  \cite{hatamizadeh2021unetr}, they studied using transfer learning (TL) for VGG-16 \cite{simonyan2014very} and IncpectionV3 \cite{szegedy2016rethinking} to detect lesions in breast US images. Another study \cite{al2019deep}, compared the performance of TL based on VGG16, ResNet \cite{he2016deep}, Inception, and NASNet \cite{zoph2018learning} for breast masses classification in US images and showed the higher performance of the NASNET comparing to the other CNNs.

Despite the success of CNNs in image processing applications, ViT models have shown superior performance to CNNs. The reason is that they lack some of the inductive biases of CNNs, such as the translation equivalence property of local convolutions. 

\subsection{Vision Transformer (ViT)}
\label{ssec:subhead}

Vision Transformers firstly introduced by Dosovitskiy \textit{et al.} \cite{dosovitskiy2020image} have shown outperforming performance comparing to SOTA CNNs in image classification applications when trained on a large scale training dataset. 
Despite the weaker inductive bias of Transformers than CNNs, these networks show competitive results with SOTA CNNs. 
Nevertheless, the high demand for large amounts of training data and computational resources limits the modification of these networks. To tackle this problem, Touvron \textit{et al.}\cite{touvron2021training} introduced a data-efficient ViT, based on the data augmentation and regularization techniques previously employed for CNNs. In addition, they improved the performance of ViT by using a Transformer-based teacher-student approach for the image classification task.

Considering the high performance of ViTs, many researchers have studied ViT models in various vision tasks \cite{yu2021mil}. In the area of object detection, Carion \textit{et al.} \cite {carion2020end} proposed a new architecture for object detection systems using a set-based global loss and a Transformer encoder-decoder algorithm and showed on par results with the dominant R-CNN method on the challenging COCO dataset.

In image segmentation, a free convolution network, purely based on the attention-based algorithms, was provided for 3D medical image segmentation \cite{karimi2021convolution}.
Another recent study \cite {hatamizadeh2021unetr} has also leveraged the Transformer's power in 3D medical image segmentation,  providing a novel U-Net Transformer architecture employing the advantages of both U-Net and Transformer networks in image segmentation and global attention feature of Transformers. In their work, a Transformer and a CNN-based architecture are used as encoder and decoder, respectively. In another study \cite {dai2021transmed}, by Dai \textit{et al.} to employ both CNN localized receptive fields for low-level feature extractions and large-scale attention of Transformers, an algorithm composed of both CNNs and Transformers has been provided to classify multi-modal images. They showed that this strategy outperforms other SOTA CNN-based networks. 

Although many efforts have been made to improve the ViT-based models, and many ViT models pre-trained on large-scale datasets are freely available, there is still a question about choosing a pre-trained model for TL. There are many choices regarding model size and different configurations to select a model and fine-tune the weights based on the new dataset.

As shown in \cite{steiner2021train}, pre-training of a ViT model can be performed under various settings on different datasets, and any of these settings results in different performances. In this regard, Steiner \textit{et al.}  provide a large number of pre-trained ViT models with different sizes and also hybrids with ResNets on the various dataset sizes, ImageNet-1k and Imagenet-21k. Their suggestion is to select a few pre-trained models with high performance when fine-tuning for an upstream task as \textit{recommended models}. Then, instead of adapting all the pre-trained Transformers, which is an extensive task, one can choose the model from these \textit{recommended models} with the best performance for further adaptation.

\section{METHODS}
\label{sec:typestyle}

The overall procedure of our work in this paper is based on the method provided in \cite{steiner2021train}, discussed in the previous section. We transferred the \textit{recommended models} pre-trained ViT models and adapted the best one based on our specific data task. 

\subsection{Dataset and evaluation metrics}
\label{ssec:subhead}

The data used in this study includes two different datasets on breast US images. The first dataset is published online by Al-Dhabyani \textit{et al.} \cite{al2020dataset}, which has 780 breast US images (referenced as BUSI), collected from 600 women with an average image size of 500 x 500 pixels. The dataset consists of 133 normal images, 437 malignant masses, and 210 benign tumors. The second dataset, considered as dataset B \cite {yap2017automated}, includes 163 images, with an average size of 760 x 570 pixels, categorized in two classes, 110 benign masses, and 53 cancerous masses. Examples of breast US images are shown in Figure \ref{fig:res1}. 
The performance metrics we use for evaluation purposes are common metrics employed in medical image classification studies \cite{lazo2020comparison}, including classification accuracy (Acc) and area under the receiver operating characteristic curve (AUC).

\begin{figure}[t!]

\begin{minipage}[b]{.3\linewidth}
  \centering
  \centerline{\includegraphics[height=2.3cm, width=2.5cm]{benign.png}}
  \centerline{(b) Benign}\medskip
\end{minipage}
\hfill
\begin{minipage}[b]{.3\linewidth}
  \centering
  \centerline{\includegraphics[height=2.3cm, width=2.5cm]{malignant.png}}
  \centerline{(b) Malignant}\medskip
\end{minipage}
\hfill
\begin{minipage}[b]{0.3\linewidth}
  \centering
  \centerline{\includegraphics[height=2.3cm, width=2.5cm]{normal.png}}
  \centerline{(c) Normal}\medskip
\end{minipage}
\caption{Example of breast US images with three different classifications.}
\label{fig:res1}
\end{figure}

\subsection{ViT architecture}
\label{ssec:subhead}
\begin{figure}[t!]

\begin{minipage}[b]{1.0\linewidth}
  \centering
  \centerline{\includegraphics[width=8.5cm]{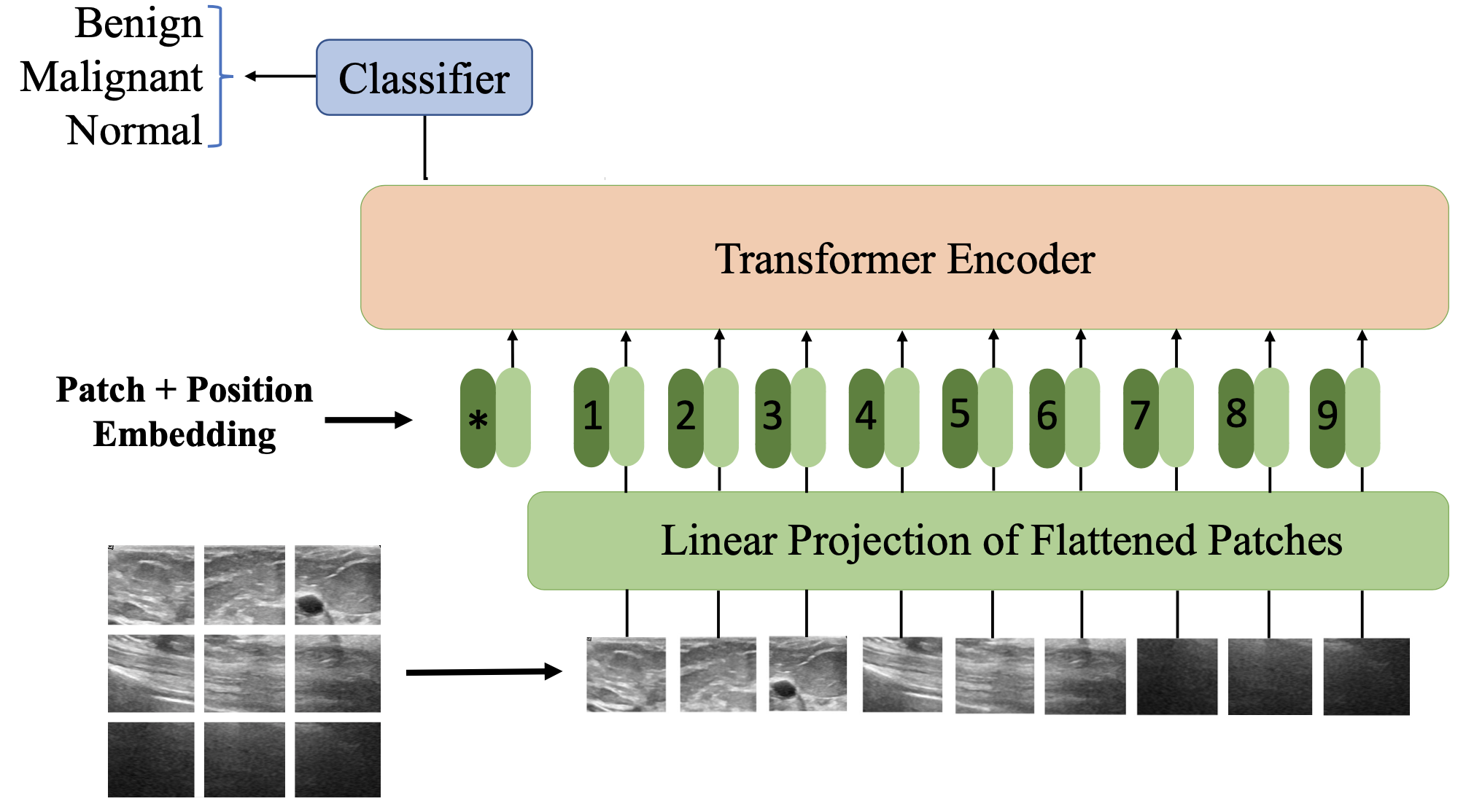}}
\end{minipage}
%
\caption{Overview of the vision Transformer used in classification of breast US.}
\label{fig:res}
\end{figure}

Following the procedure suggested in \cite {steiner2021train}, the established architecture is the same as the original ViT design introduced by \cite{dosovitskiy2020image}, except for substituting the MLP head with a linear classifier. An overview of the ViT design architecture is presented in Figure \ref{fig:res}.
In summary,  the input image is split into patches in a ViT model. A sequence of 1D patch embeddings is fed to the Transformer encoder, where self-attention modules are utilized to calculate the relation-based weighted sum of the outputs of each hidden layer. Consequently, this strategy allows the Transformers to learn global dependencies in the input images. 
\subsection{Models}
\label{ssec:subhead}

 For transfer learning of CNNs, we use the primarily pre-trained CNN networks in classification, including VGG16 \cite{simonyan2014very}, ResNet50 \cite{he2016deep}, InceptionV3 \cite{szegedy2016rethinking}, and NASNetLarge \cite{zoph2018learning} based on the breast US dataset. Consequently, we compare the CNN-based and ViT-based architectures results.
 
Furthermore, the TL approach for Transformer-based models is based on the \textit{recommended models} provided by \cite {steiner2021train}, including the main ViT models with different sizes from \cite {dosovitskiy2020image, touvron2021training} (ViT-Ti, ViT-S, ViT-B), and also the hybrid ViT+ResNet models (R+Ti, R26+S) from \cite {steiner2021train}, with the specifications provided in Table 1 and Table 2 of \cite {steiner2021train}. In these models Ti, S, B are the representations of tiny, small and based models \cite{dosovitskiy2020image}, and the models including R, represents the ViT+ResNet hybrid models.

\subsection{Fine-tuning details}
\label{ssec:subhead}

We perform a three-classes classification task of benign, malignant or normal. We split all datasets to 85\% and 15\% for training and testing, respectively. All experiments are conducted on the fixed training and testing dataset for comparison purposes. The 5-fold cross-validation (CV) is used for the final evaluation.

In CNN TL experiments, the output layer is altered with a classifier with the softmax activation function. The optimizer used is the Adam, and the models are trained for 30 epochs (with early stopping).

For ViT models, fine-tuning is performed based on the code provided in \cite {steiner2021train}, with some pre-processing modifications based on the unique physical characteristics of the US images. The loss function we use is a weighted cross-entropy loss to handle our imbalanced training dataset, optimized by stochastic gradient descent (SGD) with 0.9 for momentum. For fine-tuning, a batch size of 128 and a cosine decay learning rate starting from 0.001 with 10 linear warmup steps are selected, and the algorithm is run for 250 steps (with early stopping).

A summary of our method is as follows: 
\begin{itemize}
 
  \item Classification of BUSI+B dataset using  TL based on CNN models. 
  \item Classification of BUSI+B dataset using  TL based on ViT-based networks.
  \item Classification of augmented BUSI+B dataset using  TL based on ViT-based networks.
  \item Classification of three B, BUSI, and BUSI+B datasets using  TL based on two ViT-based and CNN networks.
\end{itemize}

\section{RESULTS}
\label{sec:majhead}

\begin{table}[t!] 
\centering
\caption{Comparing the performance of CNN models using TL on the classification of BUSI+B dataset (best values in bold font).}
\vspace*{1mm}
\centering
\begin{tabular}{c|c c c c c} 
\hline\hline
 Evaluation &ResNet & VGG & Inception & NASNET\\ [0.5ex]
 \hline 
  Acc& \textbf{85.3\%} & 82\% & 80\% & 79\% \\ 

 AUC & \textbf{0.94} & 0.92 & 0.92 & 0.917  \\

\end{tabular}
\label{table1}
\end{table}
\begin{table}[t!] 
\caption{Comparing the performance of \textit{recommended} ViT Models using TL on the classification of BUSI+B dataset (best values in bold font).}
\vspace*{1mm}
\centering
\begin{tabular}{c|c c c c c} 
\hline\hline
 Evaluation & R+Ti/16 & S/32 & B/32 & Ti/16 & R26+S/16\\ [0.5ex]
 \hline
 Acc & 85.7\% & 86\% &\textbf{ 86.7}\% & 85\% & 86.4\% \\ 
 AUC & 0.94 & 0.95 & \textbf{0.95} & 0.94 & 0.95 \\ 

\end{tabular}
\label{table2}
\end{table}

We tested the CNNs and ViT models for transferring pre-trained models to classify breast US images into three categories: benign, malignant, and normal. 

\subsection{Convolutional neural network (CNN) models}
\label{ssec:subhead}

The classification accuracy (Acc) and AUC results for different CNN models on the breast US dataset are reported in Table \ref{table1}. This table demonstrates that among different SOTA CNN networks, including ResNet50 \cite{he2016deep}, VGG16 \cite{simonyan2014very}, Inception \cite{szegedy2016rethinking}, and NASNET \cite{zoph2018learning}, employing TL based on the ResNet50 model has the best results with 85.3\% Acc and 0.95 AUC.    

\subsection{ViT-based models}
\label{ssec:subhead}
Table \ref{table2} shows the classification results for ViT models using TL. 
 
Firstly, this table illustrates the achievement of transferring pre-trained self-attention based models on breast US images classification with more than 85\% Acc for all ViT models. The best accuracy and AUC are 86.7\% and 0.95 for B/32 model, respectively. 

The classification performance obtained from the TL of ViT models (Table  \ref{table2}) and CNN models (Table \ref{table1}) shows comparable or even better results for ViT models than the corresponding results for the SOTA CNN networks. According to these tables, the best results of CNN networks (ResNet) are comparable with the results of ViT models, whereas for the other CNN models (VGG, Inception, and NASNET), the ViT networks have a better performance. These findings indicate the representation power of attention-based models in the area of US images analysis.

The interesting point in our results is the small size of the dataset used for pre-training compared to the larger size of dataset used for fine-tuning in \cite {steiner2021train}, which shows the potential power of attention-based models in medical US images analysis. The possible reason for the effectiveness of ViT models on such a small US dataset might be that, unlike natural images, the relation between spatial information or more specifically the large-scale dependencies between different patches are much more explicit in US images.

The important observation is that the results for different ViT architectures are almost the same, about 86\% Acc and 0.95 corresponding AUC. This similarity between different model outcomes indicates that  a smaller model with fewer number of parameters would be more beneficial in terms of fine-tuning duration and the computational cost.
Therefore, we choose B/32, a relatively small ViT architecture with the best Acc and AUC in the rest of our experiments.

\subsection{Data-augmentation}
\label{ssec:subhead}
In the next step, we examined the effect of using data augmentation techniques on the classification performance when fine-tuning the ViT models. Due to the special physical characteristics of US images, the augmentation techniques that can be applied are limited. In particular, we use light cropping, rotation, brightness, and contrast. As presented in Table  \ref{table3}, in consistence with the results presented in Figure 1 of \cite {steiner2021train}, augmentation does not make any improvement in transferring pre-trained ViT models.

\begin{table}[t!] 
\caption{Comparing the performance of \textit{recommended} ViT Models using TL on the classification of BUSI+B dataset with augmentation.}
\vspace*{1mm}
\centering
\begin{tabular}{c|c c c c c} 
\hline\hline
 Evaluation & R+Ti/16 & S/32 & B/32 & Ti/16 & R26+S/16\\ [0.5ex]
 \hline
 Acc & 82\% & 82\% & 81\% & 78\% & 80\% \\ 
 AUC & 0.92 & 0.92 & 0.91 & 0.9 & 0.91 \\ 

\end{tabular}
\label{table3}
\end{table}

\subsection{Breast US datasets}
\label{ssec:subhead}

We also tested the B/32 and Resnet50 models on the classification of different datasets: dataset B, BUSI, and BISI+B. The results are mentioned in Table  \ref{table4}, in which for both models, the Acc and AUC or dataset BUSI+B outperform the corresponding results for datasets BUSI and B. 

\begin{table}[t!] 
\caption{Comparing the performance of B/32 andResNet Models using TL on the classification of BUSI, B, and BUSI+B datasets (best values in bold font).}
\vspace*{1mm}
\centering
\begin{tabular}{c|c|c c } 
\hline\hline
 Dataset & Evaluation & B/32 & ResNet\\ [0.5ex]
 \hline
 B & Acc & 74\% & 79\%  \\ 
     &AUC & 0.8 & 0.84\\ [0.2cm]
 BUSI & Acc & 82\% & 83\%  \\ 
     &AUC & 0.91& 0.92\\ [0.2cm]
 BUSI+B & Acc & \textbf{86.7}\% & \textbf{85.3}\%  \\ 
     &AUC & \textbf{0.95} & \textbf{0.94}\\

\end{tabular}
\label{table4}
\end{table}

Based on Table \ref{table4}, the best results for both networks are observed when both datasets, BUSI+B, are used for fine-tuning. This is because more training data leads to a better generalization of the model. Also, the difference between the results of BUSI and BUSI+B is much less than that of datasets B and BUSI+B, considering the smaller size of dataset B with just two categories of breast US images, malignant vs. benign.

\section{CONCLUSION}
\label{sec:print}

In this study, we apply Transformer-based structures to breast US images for the breast lesion classification task for the first time. Since ViT models require a large-scale dataset for training and the size of medical imaging is relatively small, we use the strategy of transferring pre-trained ViT models for further adaptation. We compare our outcomes with the corresponding performance of the SOTA CNNs and show that attention based ViT models achieve comparable performance with CNN methods. This study shows the potential of ViT models in US image classification and, therefore, the effectiveness of learning the global anatomical dependencies of US medical images. The results presented in this study open new windows for using self-attention based architectures as an alternative to CNNs in various US medical image analysis tasks. 

\section{Compliance with ethical standards}
\label{sec:ethics}

Ethical approval was *not* required as this research study was conducted retrospectively using human subject data made available in an open access article \cite{al2020dataset}, under the CC BY license (http:// creativecommons.org/licenses/by/4.0/).

\section{Acknowledgments}
\label{sec:acknowledgments}
We acknowledge the support of the Natural Sciences and Engineering Research Council of Canada (NSERC), and thank NVIDIA for the donation of the GPU.
\bibliographystyle{IEEEbib}
\bibliography{strings,refs}

\end{document}